# RegNet: Multimodal Sensor Registration Using Deep Neural Networks

Nick Schneider[1,2], Florian Piewak[1,2], Christoph Stiller[2] and Uwe Franke[1]

***Abstract*— In this paper, we present *RegNet*, the first deep convolutional neural network (CNN) to infer a 6 degrees of freedom (DOF) extrinsic calibration between multimodal sensors, exemplified using a scanning LiDAR and a monocular camera. Compared to existing approaches, *RegNet* casts all three conventional calibration steps (feature extraction, feature matching and global regression) into a single real-time capable CNN. Our method does not require any human interaction and bridges the gap between classical offline and target-less online calibration approaches as it provides both a stable initial estimation as well as a continuous online correction of the extrinsic parameters. During training we randomly decalibrate our system in order to train *RegNet* to infer the correspondence between projected depth measurements and RGB image and finally regress the extrinsic calibration. Additionally, with an iterative execution of multiple CNNs, that are trained on different magnitudes of decalibration, our approach compares favorably to state-of-the-art methods in terms of a mean calibration error of $0.28°$ for the rotational and $6\,\mathrm{cm}$ for the translation components even for large decalibrations up to $1.5\,\mathrm{m}$ and $20°$.**

## I. INTRODUCTION

To acquire a redundant and powerful system for autonomous driving, recent developments rely on a variety of optical sensors. Especially the fusion of camera and depth sensors has therefore been studied intensively in the last few years. To combine the information of those sensors, a common world coordinate system has to be defined in respect to which the sensors' poses are given. Transforming a point $\boldsymbol{x}$ given in the sensor coordinate system into a point $\boldsymbol{y}$ in the world coordinate system is typically modeled via an affine transformation matrix $\boldsymbol{H}$, i.e.

$$\boldsymbol{y} = \boldsymbol{H}\boldsymbol{x} \; . \tag{1}$$

The task of estimating the transformation matrix $\boldsymbol{H}$ is called *extrinsic calibration* and has been studied for a variety of sensor modalities and combinations. Most approaches can be divided into three steps:

1) Find distinct features in the sensor data, e.g. corners or artificial targets;
2) Use those features to establish correspondences between the sensors;
3) Given the correspondences, determine $\boldsymbol{H}$ by solving a system of equations or by minimizing an error function.

The extraction of distinct features can be challenging as correspondences have to be made across different sensor modalities. Most offline calibration approaches therefore rely on special calibration targets which provide strong and distinct signals in all modalities, allowing for an easy detection and localization[1][2][3]. However, those approaches are time consuming as they need human interaction for feature selection or they have to be performed in a controlled environment. Therefore, several online calibration methods have been proposed recently[4][5][6][7]. The challenging part in online calibration is to find matching patterns in an unstructured environment. Most of the state-of-the-art approaches do so by using handcrafted features such as image edges. Because the descriptors of such features are often not discriminative, the matching fails, and the subsequent optimization does not lead to satisfying results; especially when facing large calibration errors.

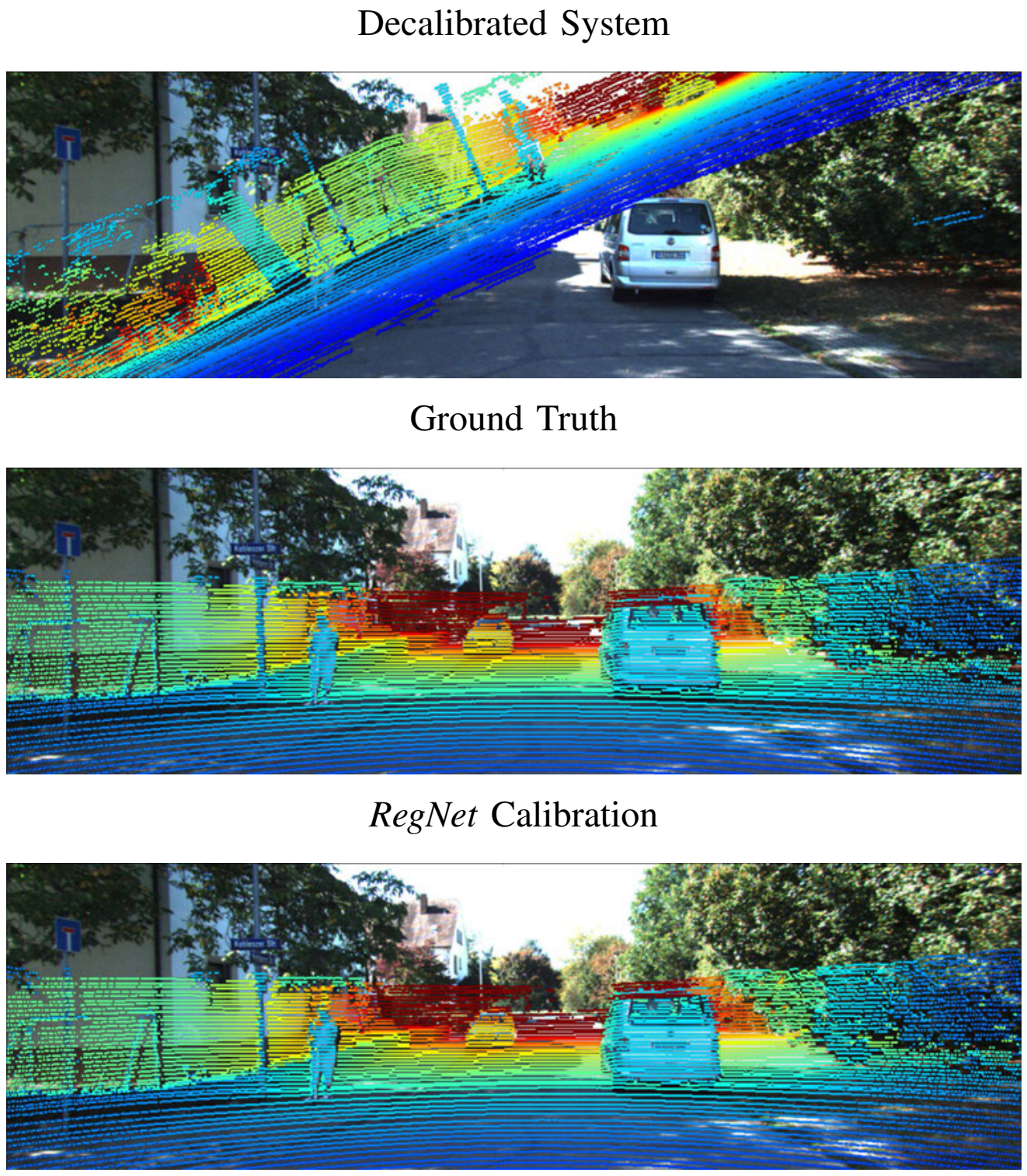

Fig. 1: *RegNet* is able to correct even large decalibrations such as depicted in the top image. The inputs for the deep neural network are an RGB image and a projected depth map. *RegNet* is able to establish correspondences between the two modalities which enables it to estimate a 6 DOF extrinsic calibration.

In this work, we present *RegNet*, the first CNN to fully regress the extrinsic calibration between sensors of different modalities. We solve the tasks of feature extraction, feature matching, and global optimization in real-time by using an end-to-end trained deep neural network. We propose a simple yet effective sampling strategy which allows us to generate an infinite amount of training data from only one manually conducted extrinsic calibration. After the network has been trained, our approach does not need any further human interaction. This is a huge advantage for e.g. series production of autonomous vehicles where only a single car has to be calibrated manually in order to train *RegNet*, which then calibrates all remaining vehicles. Furthermore, the network is able to monitor and correct calibration errors online, without the need of returning to a controlled environment for recalibration.

[1]Daimler AG, R&D, Böblingen, Germany
[2]Karlsruhe Institute of Technology, Karlsruhe, Germany
Primary contact: `nick.schneider@daimler.com`

## II. RELATED WORK

Recently, many multi-sensor indoor [8], [9] and outdoor [10], [11] datasets have been released, encouraging the research community to advance the state-of-the-art in various scene understanding tasks by exploiting multi-modal input data. Fusing sensor data on a low-level requires a highly accurate registration of the various sensors. Therefore, extrinsic calibration is an important field of research; especially the registration of sensors with different modalities is challenging. In this work, we focus on the calibration of camera and depth sensors due to their relevance in the field of autonomous driving. Most state-of-the-art approaches handle the 3D-2D registration between a camera and a depth sensor by using special calibration targets [1], [2], [3]. Other semi-automatic methods extract human- selected 3D and 2D shapes from both sensors which are then aligned [12], [13]. The mentioned methods achieve excellent results and can therefore be used for a suitable initial calibration. However, they are either time consuming [12], [13], [2], [3]. or require a controlled environment [1].

Once a sensor system goes online and starts to operate, e.g. as a product or test fleet vehicle, external forces such as mechanical vibrations or temperature changes may decrease the calibration quality. In this case, the system has to detect and correct such decalibrations. This is referred to as online calibration and has been investigated in several recent studies. In [4] for example LiDAR scans are aligned to camera images by matching projected depth edges to image edges. A similar approach is proposed by Levinson et al. [5]. They calculate depth gradients on a LiDAR point cloud and project those gradients onto an inverse distance transform of the edge image. If strong gradients are associated to pixels which are close to an image edge this results in a low energy. The subsequent optimization determines the calibration parameters by energy minimization. In a more recent work, Pandey et al. [6] realize a LiDAR-camera calibration by means of mutual information maximization. The mutual information is computed by comparing the intensity readings of the projected LiDAR points with the camera intensities. Chien et al. [7] identified weaknesses of the aforementioned approaches especially at highly textured surfaces and shadows, which were wrongly used as targets. Furthermore, the approaches could not deal with occlusions due to the sensor displacements. Therefore, a visual-odometry driven online calibration is proposed. They argue that the performance of the estimated ego-motion is directly correlated to the quality of the extrinsic parameters. As the correct ego-motion is unknown in their experiments, they evaluate the inverse re-projection error function. As the smoothness and convexity of this function was not sufficient for a robust energy minimization they added constraints using the approaches of Levinson et al. [5] and Pandey et al. [6]. The combination leads to stable results if the calibration is disturbed by not more than $2^\circ$ and $10\,\mathrm{cm}$. However, we experienced that the energy minimization often gets stuck in local minima which is why the approach cannot compensate for larger errors. Furthermore, the approach is not real-time capable as it solves for the visual odometry and extrinsic parameters iteratively using the Levenberg-Marquardt algorithm and gradient-descent.

At the same time, deep learning has been successfully applied to classic computer vision tasks such as optical flow [14], [15] or stereo estimation [16], [17]. Kendall et al. [18] train a network to regress a 6 degree of freedom (DOF) camera pose. Ummenhofer et al. [19] combine elements of the aforementioned and estimate depth, flow, and ego-motion to calculate structure-from-motion using an end-to-end trained deep neural network. Surprisingly, there are only few works leveraging the strength of deep learning for calibration. Workman et al. [20] estimate the focal length of their system given natural images whereas Giering et al. [21] use multi-modal CNNs for real- time LiDAR-video registration. By concatenating flow, RGB and LiDAR depth patches they solve a 9-class classification problem where each class corresponds to a particular $x$-$y$ shift on an ellipse. However, to the best of our knowledge, there exists no deep learning-based approach that directly regresses the calibration parameters.

In this work, we leverage the strength of deep neural networks for feature extraction, feature matching, and regression to estimate the extrinsic calibration parameters of a multi-modal sensor system. To this end, we propose *RegNet* based on our main contributions:

1) A CNN that directly regresses extrinsic calibration parameters for all 6 DOF;
2) an effective training strategy, which needs only one manually calibrated sensor setup;
3) a real-time, low-memory network, which can be easily deployed in autonomous vehicles.

## III. METHOD

The goal of this work is to develop a generic approach for extrinsic sensor calibration. For this purpose we leverage deep neural networks for feature extraction and matching like proposed by Dosovitskiy et al. [14] and Ilg et al. [15] and regress a full 6 DOF extrinsic calibration which is motivated by the work of Kendall et al. [18].

Although our approach can generally be applied to different sensor modalities and combinations, we focus on LiDAR-camera calibration in this work due to their important role in autonomous driving. In the following sections we discuss the data representation of the network inputs, the design of the CNN, subsequent refinement steps, and training details.

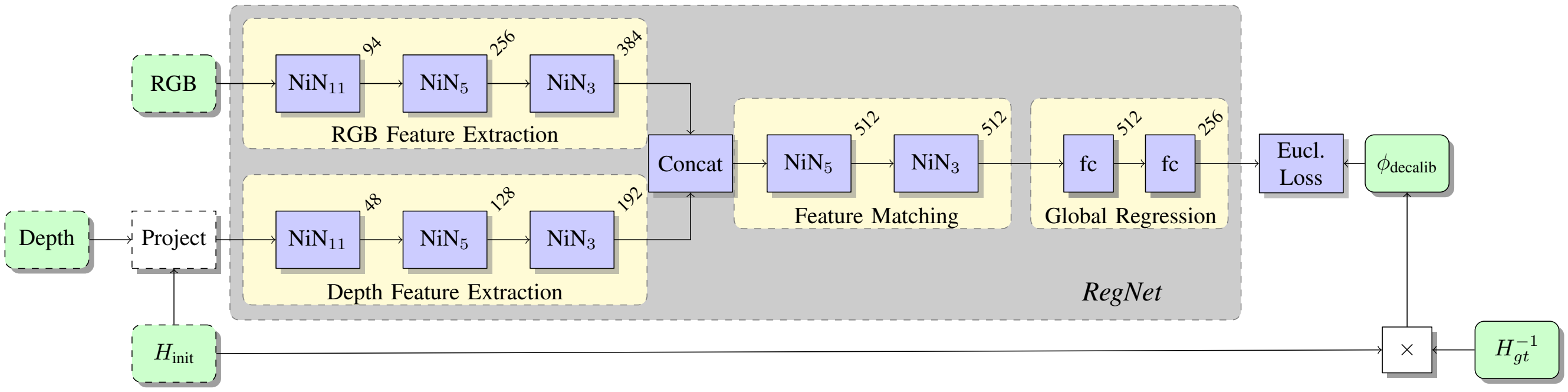

Fig. 2: Our method estimates the calibration between a depth and an RGB sensor. The depth points are projected on the RGB image using an initial calibration $\boldsymbol{H}_{\text{init}}$. In the first and second part of the network we use NiN blocks to extract rich features for matching. The kernel size $k$ of the first convolutional layer of the NiN block is displayed by the indices. The number of feature channels is shown in the top right corner of each layer module. The final part regresses the decalibration by gathering global information using two fully connected layers. During training $\phi_{\text{decalib}}$ is randomly permutated resulting in different projections of the depth points.

### A. Data Collection and Representation

The performance of deep learning methods improve with the accuracy and amount of data presented to them. In our case we would need pairs of images and LiDAR scans accompanied with a ground truth calibration. However, determining the ground truth for thousands of differently arranged LiDAR-camera pairs would be bothersome. We therefore reformulate the problem of extrinsic calibration as determining the decalibration $\phi_{\text{decalib}}$ given an initial calibration $\boldsymbol{H}_{\text{init}}$ and a ground truth calibration $\boldsymbol{H}_{\text{gt}}$. We can then vary $\boldsymbol{H}_{\text{init}}$ randomly to get a huge amount of training data.

To be able to establish correspondences, the LiDAR points are projected on the camera frame using $\boldsymbol{H}_{\text{init}}$ and the intrinsic camera matrix $\boldsymbol{P}$, i.e.

$$z_c \begin{bmatrix} u \\ v \\ 1 \end{bmatrix} = \boldsymbol{P} \ \boldsymbol{H}_{\text{init}} \ \boldsymbol{x} \ . \tag{2}$$

At each pixel $(u, v)$ we store the inverse depth of the projected point (in camera coordinates) $z_c$ or zero if no LiDAR point was projected on that particular pixel. As most common LiDAR sensors provide only few measurements in comparison to the amount of image pixels, the depth images are quite sparse. To account for the sparsity, we upsample the projected LiDAR points by using max pooling on the input depth map. The LiDAR depth image as well as the camera image are mean adjusted.

The ground truth decalibration can be represented in various ways. The homogeneous decalibration matrix $\phi_{\text{decalib}}$ is composed of a $3 \times 3$ rotation matrix $\boldsymbol{R}$ and a $3 \times 1$ translation vector $\boldsymbol{t}$:

$$\phi_{\text{decalib}} = \begin{bmatrix} & \boldsymbol{R} & & \boldsymbol{t} \\ 0 & 0 & 0 & 1 \end{bmatrix} \tag{3}$$

To reduce the amount of learned parameters, the rotation can be represented by Euler angles. Another option would be to use quaternions like proposed in [18]. However, Euler angles and quaternions share the disadvantage of decoupled rotation and translation parameters. With dual-quaternions a unified representation of translation and rotation can be achieved. A dual-quaternion $\boldsymbol{\sigma}$ is composed of a real part $\boldsymbol{p}$ and a dual part $\boldsymbol{q}$:

$$\boldsymbol{\sigma} = \boldsymbol{p} + \epsilon \boldsymbol{q} \tag{4}$$

where $\boldsymbol{p}$ contains rotational and $\boldsymbol{q}$ rotational and translational information. The decalibration values are normalized to the range $[-1, 1]$. For quaternions we can use the normalized form with $\|\boldsymbol{p}\| = 1$. For dual quaternions, $\boldsymbol{q}$ is represented with values without a specific range. This results in an imbalance of the dual quaternions during training. To compensate this effect, we multiply the values of $\boldsymbol{p}$ by a factor $f$. This also creates an implicit weighting of the rotational part for the loss function of the CNN.

### B. Network Architecture

We design our network to solve the tasks of feature extraction, feature matching and regression of the calibration parameters. All three steps are combined in only one CNN which can be trained end-to-end. The block diagram in Figure 2 shows the outline of *RegNet* and it's embedding in the training pipeline.

Due to their fast convergence we constructed the network by arranging several Network in Network (NiN) blocks which have been proposed by Lin et al. [22]. A NiN block is composed of one $k \times k$ convolution followed by several $1 \times 1$ convolutions.

**Feature Extraction.** We encourage the network to extract a rich feature representation for each modality individually. Therefore, we first process the RGB and LiDAR depth map separately, resulting in two parallel data network streams. For the RGB part we use the weights and architecture proposed by Lin et al. for ImageNet [23] classification. However, we skip the last NiN block as we're only interested in the feature extraction part and not in image classification. The depth stream is kept symmetrically but with a fewer number of feature channels as this part is learned from scratch.

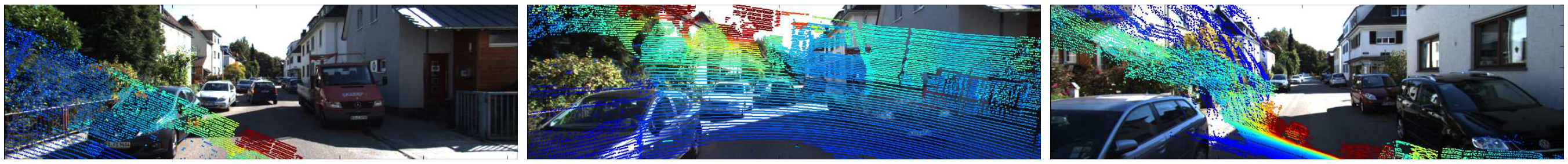

Fig. 3: We deviate the initial calibration up to $20°$ in rotation and up to $1.5\,\mathrm{m}$ in translation from the ground truth calibration. This might result in projections of the LiDAR points where most of the points are outside the image area and it is therefore difficult to establish correspondences with the RGB image.

**Feature Matching.** After extracting features from both input modalities the feature maps are concatenated to fuse the information from both modalities. This part of the network is also realized as a stack of NiN blocks. By convolving the stacked LiDAR and RGB features a joint representation is generated. This architecture was motivated by Dosovitskiy et al. [14] who also introduced a specific correlation layer. However, they show that their network is capable of correlating features without explicitly demanding it.

**Global Regression.** To regress the calibration, the global information that has been extracted from both modalities has to be pooled. This step is comparable to a global optimization or solver as used in classical calibration algorithms. To realize a global information fusion we stack two fully connected layers followed by a Euclidean loss function. Like [18] we also experienced that branching the network to handle translational and rotational components separately worsened the result.

### C. Refinement

**Iterative Refinement.** The projection of the depth points strongly varies with the given initial calibration as depicted in Figure 3. Some transformations cause the projection to be mostly outside the image area, so only few correspondences between the LiDAR and the RGB image can be established. We noted, that our network is still able to improve the calibration in those cases. By using the new estimated calibration $\hat{\boldsymbol{H}} = \boldsymbol{H}_{\text{init}} \hat{\phi}^{-1}_{\text{decalib}}$ we can again project the depth points resulting in more depth points for correlation. This step can then be iterated several times.

**Temporal Filtering.** There are only few scenarios where an instant registration between two modalities is required. In the context of autonomous driving the extrinsic calibration between different sensors might involve more than just one frame. If the output of the network is analyzed over time by using a moving average, the approach yields more robust results.

### D. Training Details

The RegNet was developed using the Caffe library introduced by [24]. After each convolutional layer but the last we add a Rectified-Linear Unit (ReLU). The training of the network is performed with the Adam solver [25]. We use Euclidean loss to infer the deviation from the ground truth decalibration. The network is trained for 3 Mio. iterations. We set the parameters of the solver to the suggested default values $\beta_1 = 0.9$, $\beta_2 = 0.999$ and $\epsilon = 10^{-8}$. The learning rate is fixed at $\alpha = 10^{-5}$ and the batch size is set to $b = 1$ - an increased batch size did not improve our results. For the RGB feature extraction part we initialize the NiN blocks with ImageNet [23]. The remaining weights are learned from scratch and are initialized using Xavier initialization [26].

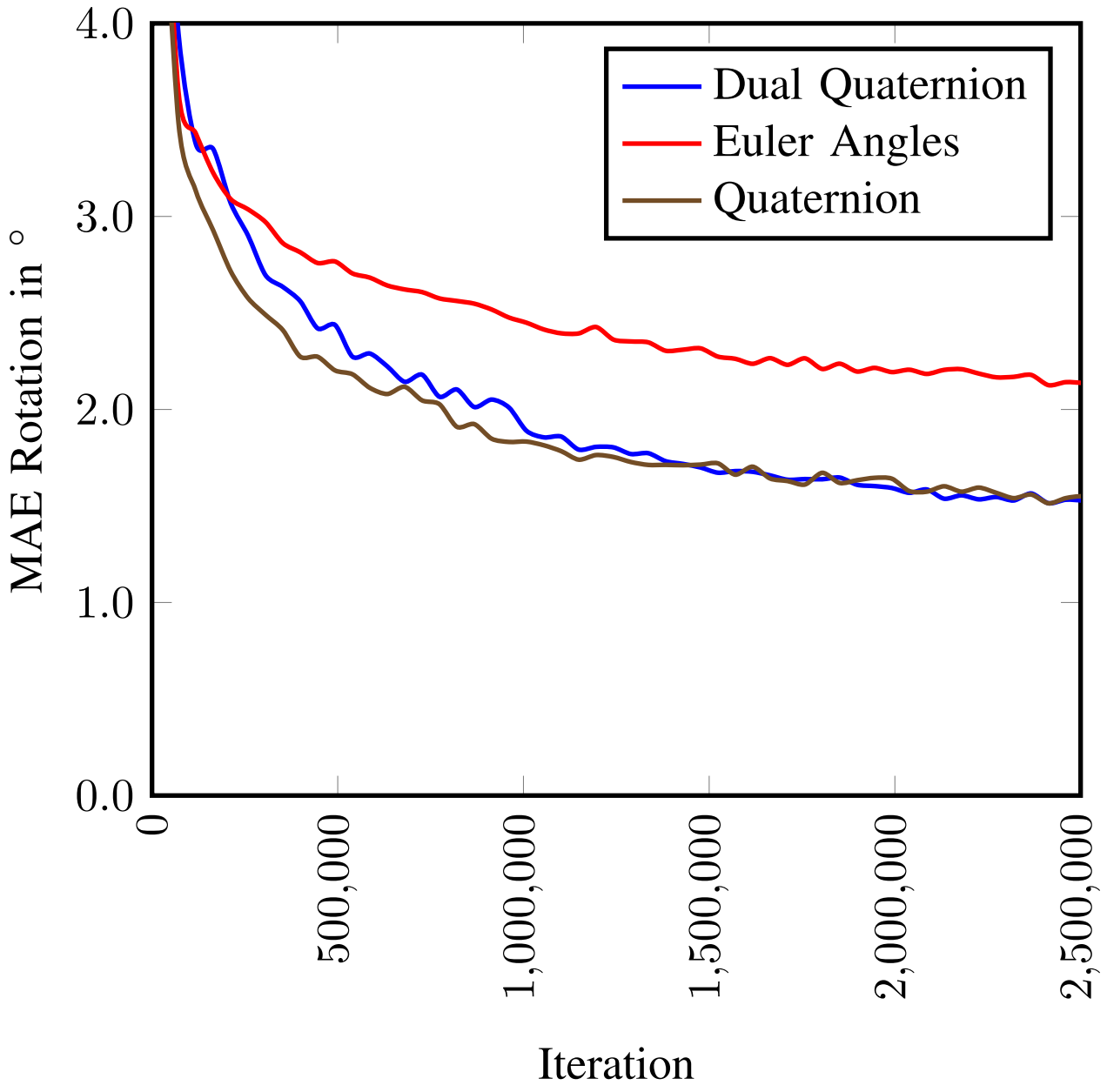

Fig. 4: Development of the mean absolute error (MAE) of the rotational components over training iteration for different output representations: Euler angles are represented in red, quaternions in brown and dual quaternions in blue. Both quaternion representations outperform the Euler angles representation.

## IV. EXPERIMENTS

To evaluate our approach, we perform several experiments on real sensor data. As we are most interested in sensors that are relevant for autonomous driving, we focus on the calibration of a LiDAR-camera setup in this section. During our experiments we noticed that the rotational components have a larger impact on the quality of the resulting registration and are also harder to determine by our network. Therefore, our comparisons between different methods are mainly based on the rotational components.

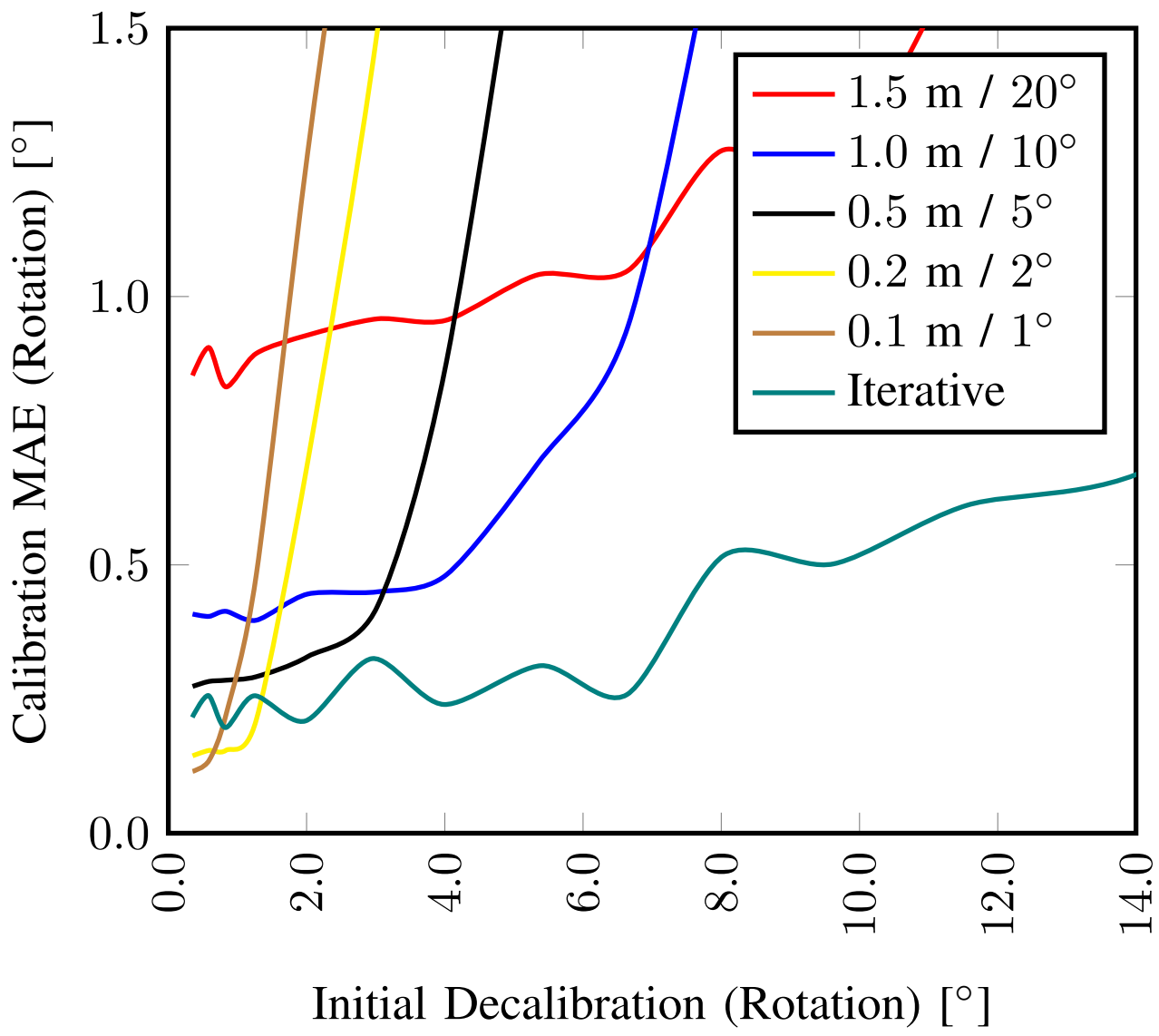

Fig. 5: Analysis of the calibration performance (rotation only) over the decalibration magnitude for different networks. The networks have been trained on random initial decalibrations, varying from 0.1m / 1° to 1.5m / 20°. It can be seen that the networks perform better on certain decalibrations, depending on the range they have been trained on. Therefore, an iterative execution of experts is proposed.

### A. Dataset

We evaluate our approach on the KITTI dataset [11], which provides $1392 \times 512$ pixel RGB images as well as depth measurements from a Velodyne HDL-64E LiDAR scanner. The extrinsic parameters of the dataset were calculated using the method of [1] and serve as ground truth for our experiments. For training, validation and testing we use the raw sequences of KITTI dataset where for each recording day different intrinsic and extrinsic calibrations were calculated. To reduce inconsistencies caused by calibration noise, we only use sequences of the first recording day (09/26/2011) for training and validation. For validation we select two challenging sequences (drive 0005 and 0070 with 574 frames in total) while all other sequences are only used for training (14863 frames). We randomly vary $\phi_{\text{decalib}}$ for each frame during training as described in Section III-A, yielding a potentially infinite amount of training data. The final testing (Section IV-D) is performed on a separate day and sequence (09/30/2011 drive 0028 with 5177 frames) to create an independent test set. We chose this sequence as it contains a huge variety of different scenes.

### B. Data Representation

The representation of the decalibration $\phi_{\text{decalib}}$ is critical for the performance of our method. In this section, we compare the results of three different representations: Euler angles with translation, quaternions with translation and dual quaternions. Each representation is normalized as described in Section III-A. We analyzed the distribution of the real part $\boldsymbol{p}$ of the dual quaternion within our decalibration range of $20°$ and determined the factor $f = 100$ to balance the dual quaternions. We also found that this factor gained the best results using quaternions with translation. Larger values of $f$ result in volatile translation whereas smaller values result in worse rotation estimates.

Figure 4 shows the mean absolute error of the estimated rotation. Both quaternion representations outperform the Euler angles. However, the curve progression of both quaternion representations suggests that dual quaternions will have a higher performance at longer training time. Subsequent experiments are therefore performed with dual quaternions only.

### C. Different Decalibration Ranges

During training we challenge the network to compensate for random decalibrations in the range of $[-1.5\,\text{m}, 1.5\,\text{m}]$ and $[-20°, 20°]$. The Euclidean loss penalizes strong deviations which is why large decalibrations have a bigger impact on the network than small ones. This results in a worse relative improvement for small decalibrations which is depicted in Figure 5. To compensate this effect, we train expert networks on different decalibration ranges. These ranges are based on the worst mean absolute error (MAE) of the network, which is trained on the next larger range, to reach high robustness. We determine the following ranges: $[-x, x]$ / $[-y, y]$ (translation / rotation) for $x = \{1.5\,\text{m}, 1.0\,\text{m}, 0.5\,\text{m}, 0.2\,\text{m}, 0.1\,\text{m}\}$ and $y = \{20°, 10°, 5°, 2°, 1°\}$.

Figure 5 shows how these expert networks perform on varying decalibration magnitudes. It can be seen that choosing the best network is dependent on the decalibration. However, as we do not know the decalibration outside of our test environment, we perform an iterative refinement as described in Section III-C starting with the $20°/1.5\,\text{m}$ network followed by the $10°/1.0\,\text{m}$, $5°/0.5\,\text{m}$, $2°/0.2\,\text{m}$ and $1°/0.1\,\text{m}$ network, respectively. A result of this iterative refinement is shown in Figure 6. The execution time of the iterative approach is real-time capable with $7.3\,\text{ms}$ for one network forward pass on an NVIDIA TITAN X (Pascal architecture).

The order of the networks is optimized for decalibration scenarios up to $1.5\,\text{m}$ and $20°$ which can be used for calibrating a sensor from scratch. In online calibration scenarios however, the decalibrations are much smaller. In this case the number of networks for iterative execution can be decreased. Figure 8 shows an online scenario of random decalibrations up to $20\,\text{cm}$ and $2°$, where only two networks with decalibration ranges $[-x, x]$ / $[-y, y]$ (translation / rotation) for $x = \{0.2\,\text{m}, 0.1\,\text{m}\}$ and $y = \{2°, 1°\}$ are executed. Within this online scenario we reach the same performance as in the offline scenario while decreasing the execution time by using only two networks iteratively.

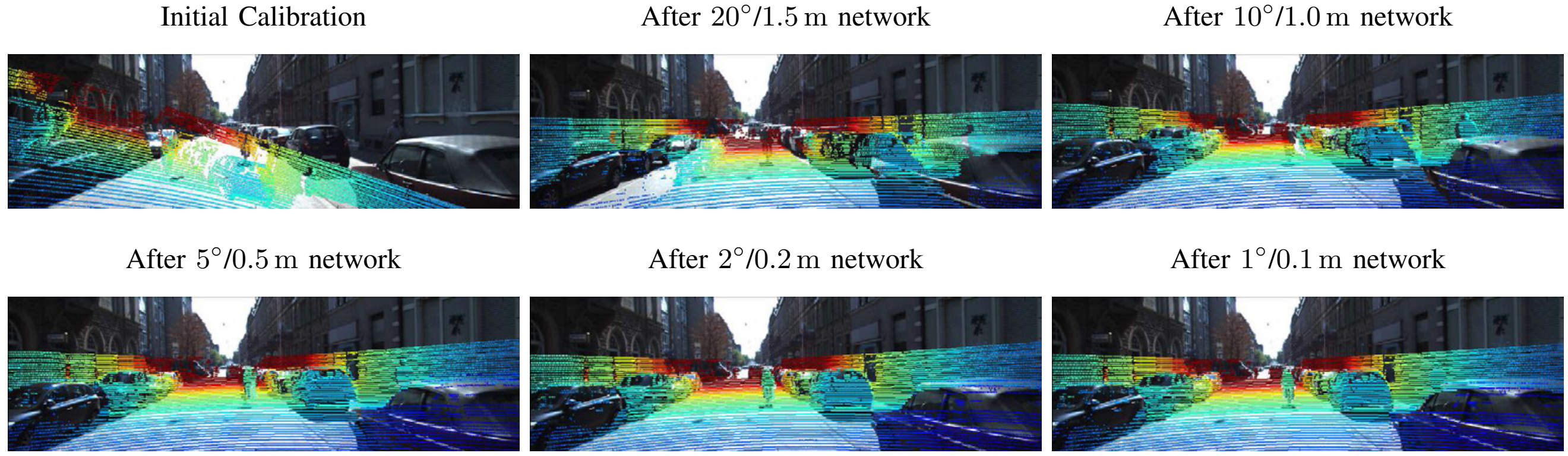

Fig. 6: For the iterative refinement, the estimated calibration of one expert network is used to improve the projection of the depth points. The refined depth map is then forwarded to the next network. From top left to bottom right we can see a constant improvement in each iteration step.

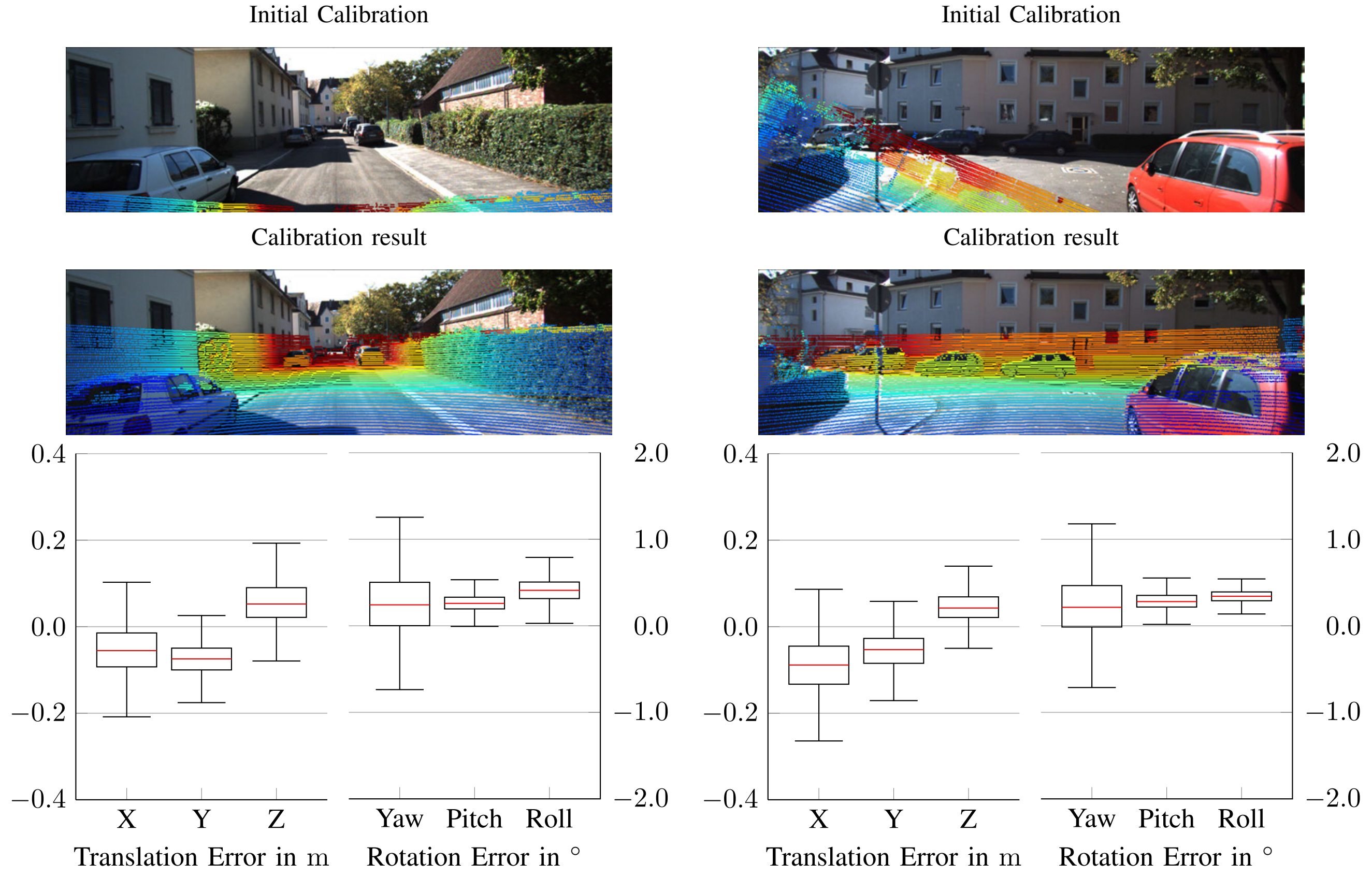

Fig. 7: Examples of the Distribution of the calibration error for a decalibration, which is fixed over the test sequence. Five networks are executed iteratively ($20°/1.5\,\mathrm{m}$, $10°/1.0\,\mathrm{m}$, $5°/0.5\,\mathrm{m}$, $2°/0.2\,\mathrm{m}$ and $1°/0.1\,\mathrm{m}$).

### D. *Temporal Filtering*

The previous experiments are based on only one frame and can be noisy due to missing structure and sensor artifacts like rolling shutter or dynamic objects. This can be further improved by analyzing the results over time as mentioned in Section III-C. For this purposes, we determine the distribution of our results over the whole test sequence, while keeping the decalibration fixed. Figure 7 visualizes two examples of the distributions of the individual components by means of boxplots. In general, the estimated decalibrations $\hat{\phi}_{\text{decalib}}$ are distributed well around the ground truth values. Taking the median over the whole sequence resulted in the best performance on the validation set. For the quantitative evaluation on the test set we sampled decalibrations in the range of $[-20°, 20°]$ / $[-1.5\,\mathrm{m}, 1.5\,\mathrm{m}]$. The decalibration is kept fixed for one pass of the test set and then resampled. In total we performed 100 runs on the test set with different decalibrations. Our approach achieves a mean angle error of $0.28°$ (yaw, pitch, roll: $0.24°$, $0.25°$, $0.36°$) and a mean translation error of $6\,\mathrm{cm}$ (x, y, z: $7\,\mathrm{cm}$, $7\,\mathrm{cm}$, $4\,\mathrm{cm}$). In Figure 1 and Figure 9 results of our approach are visualized. It can be seen that the network is capable of handling even large decalibrations from the ground truth.

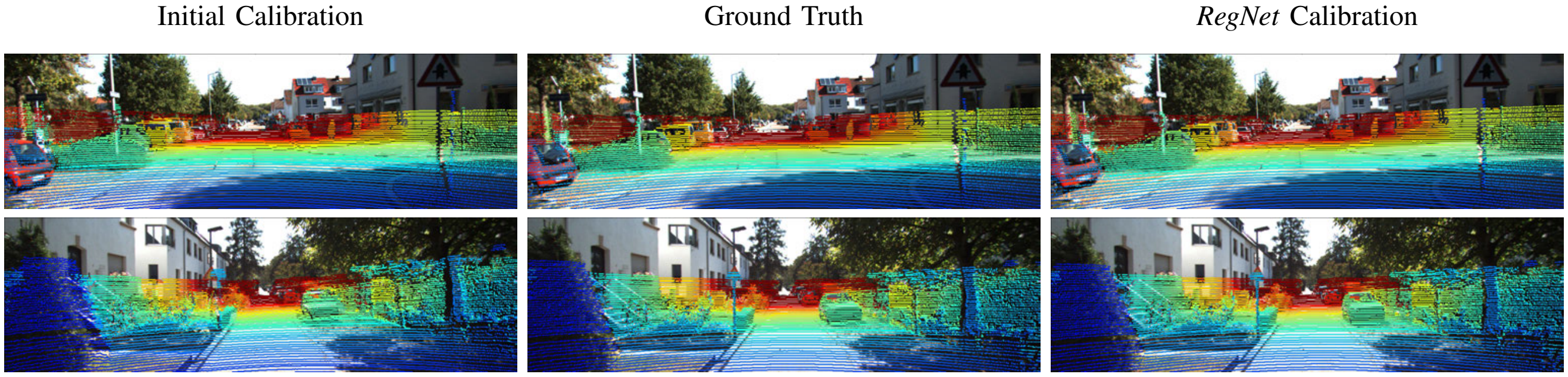

Fig. 8: Examples of calibration results for an online scenario based on decalibrations up to $0.2m$ and $2°$, where only two networks are executed iteratively ($2°/0.2\,\mathrm{m}$ and $1°/0.1\,\mathrm{m}$ network)

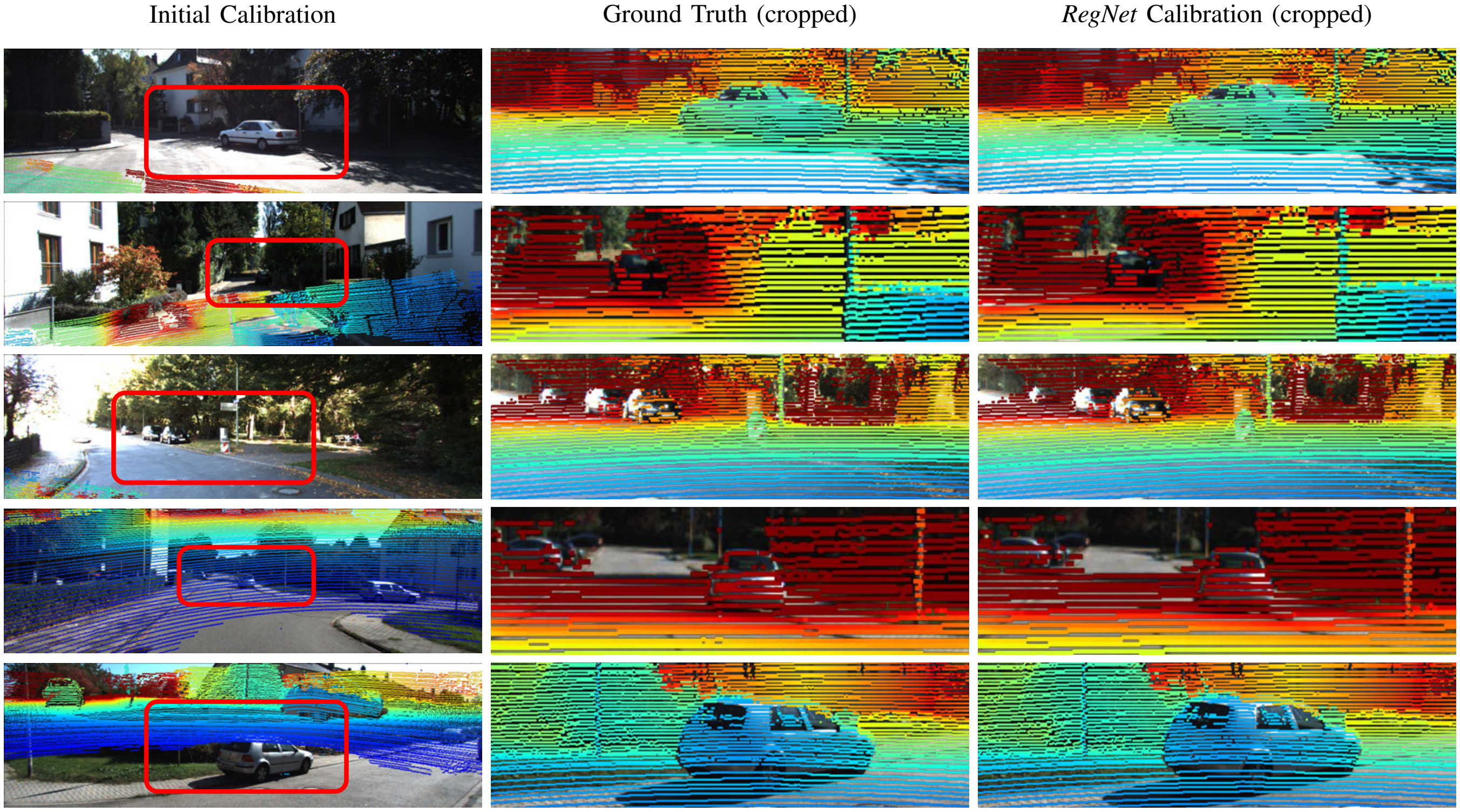

Fig. 9: Results of different single shot calibration results on the test set. Five networks, trained on different decalibration ranges ($20°/1.5\,\mathrm{m}$, $10°/1.0\,\mathrm{m}$, $5°/0.5\,\mathrm{m}$, $2°/0.2\,\mathrm{m}$ and $1°/0.1\,\mathrm{m}$), are executed iteratively. Although the initial calibration is extremely bad, the proposed method delivers accurate results.

## V. CONCLUSION

In this paper we introduced a novel approach for extrinsic calibration of multimodal sensors based on a deep convolutional neural network. Compared to existing approaches, our network concept replaces all three conventional calibration steps (feature extraction, feature matching and global regression) and directly infers the 6 DOF of the calibration. We train several networks on different decalibration ranges to iteratively refine the calibration output. With this approach different calibration tasks can be solved: on the one hand, a target-less calibration can be applied from scratch and without human interaction by using temporal filtering to reduce noise and reject outliers of a whole sequence - on the other hand online calibration can be achieved by applying a moving average or sliding window filter to adapt the calibration in real-time. Our method yields a mean calibration error of $6\,\mathrm{cm}$ for translation and $0.28°$ for rotation with decalibration magnitudes of up to $1.5\,\mathrm{m}$ and $20°$, which competes with state-of-the-art online and offline methods.

Our approach could still be improved by replacing the iterative refinement with an end-to-end trained recurrent network. This could increase the performance by optimizing the calibration ranges implicitly at training time.

# References

[1] A. Geiger, F. Moosmann, O. Car, and B. Schuster, "Automatic camera and range sensor calibration using a single shot," in *ICRA*, 2012.

[2] F. M. Mirzaei, D. G. Kottas, and S. I. Roumeliotis, "3d lidar–camera intrinsic and extrinsic calibration: Identifiability and analytical least-squares-based initialization," in *IJRR*, 2012.

[3] Q. Zhang and R. Pless, "Extrinsic calibration of a camera and laser range finder (improves camera calibration)," in *IROS*, 2004.

[4] S. Bileschi, "Fully automatic calibration of lidar and video streams from a vehicle," in *ICCV Workshops*, 2009.

[5] J. Levinson and S. Thrun, "Automatic online calibration of cameras and lasers," in *Robotics Science Systems*, 2013.

[6] G. Pandey, J. R. McBride, S. Savarese, and R. M. Eustice, "Automatic extrinsic calibration of vision and lidar by maximizing mutual information," in *Journal of Field Robotics*, 2015.

[7] H. Chien, R. Klette, N. Schneider, and U. Franke, "Visual odometry driven online calibration for monocular lidar-camera systems," in *ICPR*, 2016.

[8] S. Song, S. Lichtenberg, and J. Xiao, "Sun rgb-d: A rgb-d scene understanding benchmark suite," in *CVPR*, 2015.

[9] N. Silberman, D. Hoiem, P. Kohli, and R. Fergus, "Indoor segmentation and support inference from rgbd images," in *ECCV*, 2012.

[10] W. Maddern, G. Pascoe, C. Linegar, and P. Newman, "1 year, 1000km: The oxford robotcar dataset," in *IJRR*, 2016.

[11] A. Geiger, P. Lenz, C. Stiller, and R. Urtasun, "Vision meets robotics: The kitti dataset," in *IJRR*, 2013.

[12] L. Tamas and Z. Kato, "Targetless calibration of a lidar - perspective camera pair," in *ICCV Workshops*, 2013.

[13] L. Tamas, R. Frohlich, and Z. Kato, "Relative pose estimation and fusion of omnidirectional and lidar cameras," in *ECCV Workshops*, 2014.

[14] P. Fischer, A. Dosovitskiy, E. Ilg, P. Häusser, C. Hazırbaş, V. Golkov, P. van der Smagt, D. Cremers, and T. Brox, "Flownet: Learning optical flow with convolutional networks," in *ICCV*, 2015.

[15] E. Ilg, N. Mayer, T. Saikia, M. Keuper, A. Dosovitskiy, and T. Brox, "Flownet 2.0: Evolution of optical flow estimation with deep networks," in *In arXiv preprint arXiv:1612.01925*, 2016.

[16] J. Zbontar and Y. LeCun, "Stereo matching by training a convolutional neural network to compare image patches," in *JMLR*, 2016.

[17] N. Mayer, E. Ilg, P. Hausser, P. Fischer, D. Cremers, A. Dosovitskiy, and T. Brox, "A large dataset to train convolutional networks for disparity, optical flow, and scene flow estimation," in *CVPR*, 2016.

[18] A. Kendall, M. Grimes, and R. Cipolla, "Posenet: A convolutional network for real-time 6-dof camera relocalization," in *ICCV*, 2015.

[19] B. Ummenhofer, H. Zhou, J. Uhrig, N. Mayer, E. Ilg, A. Dosovitskiy, and T. Brox, "Demon: Depth and motion network for learning monocular stereo," in *In arXiv preprint arXiv:1612.02401*, 2016.

[20] S. Workman, C. Greenwell, M. Zhai, R. Baltenberger, and N. Jacobs, "Deepfocal: A method for direct focal length estimation s," in *ICIP*, 2015.

[21] M. Giering, V. Venugopalan, and K. Reddy, "Multi-modal sensor registration for vehicle perception via deep neural networks," in *High Performance Extreme Computing Conference*, 2015.

[22] M. Lin, Q. Chen, and S. Yan, "Network in network," in *ICLR*, 2014.

[23] J. Deng, W. Dong, R. Socher, L.-J. Li, K. Li, and L. Fei-Fei, "Imagenet: A large-scale hierarchical image database," in *CVPR*, 2009.

[24] Y. Jia, E. Shelhamer, J. Donahue, S. Karayev, J. Long, R. Girshick, S. Guadarrama, and T. Darrell, "Caffe: Convolutional architecture for fast feature embedding," in *ACM international conference on Multimedia*, 2014.

[25] D. Kingma and J. Ba, "Adam: A method for stochastic optimization," in *ICLR*, 2015.

[26] X. Glorot and Y. Bengio, "Understanding the difficulty of training deep feedforward neural networks," in *Aistats (Vol. 9, pp.)*, 2010.